\documentclass[conference]{IEEEtran}
\IEEEoverridecommandlockouts
\usepackage{cite}
\usepackage{amsmath,amssymb,amsfonts}
\usepackage{graphicx}
\usepackage{textcomp}
\usepackage{xcolor}
\usepackage{amsmath, amssymb}
\usepackage{algorithm}
\usepackage{algpseudocode}


\usepackage{geometry}
\usepackage{physics}
\usepackage{float} 
\usepackage{subcaption}
\usepackage{booktabs}
\usepackage{fancyhdr}
\usepackage{multirow}
\usepackage{wrapfig} 
\usepackage{array}    
\usepackage{booktabs} 
\usepackage{caption}  
\usepackage{adjustbox}

\def\BibTeX{{\rm B\kern-.05em{\sc i\kern-.025em b}\kern-.08em
    T\kern-.1667em\lower.7ex\hbox{E}\kern-.125emX}}

\begin{document}

\title{Federated Physics-Grounded Reinforcement Learning for Distributed Stability Control in Smart Grids}


\author{
\IEEEauthorblockN{Omar Al-Refai,
Ibrahim Shahbaz,
Adam Ali Husseinat,
Eman Hammad}
\IEEEauthorblockA{\textit{iSTAR Lab, Texas A\&M University}\\
College Station, TX, USA\\
\{omaralrefai, i.shahbaz, adamali, eman.hammad\}@tamu.edu}
}

\pagestyle{fancy}
\fancyhf{}
\renewcommand{\headrulewidth}{0pt}
\renewcommand{\footrulewidth}{0pt}
\fancyfoot[C]{\scriptsize © 2026 IEEE. Personal use of this material is permitted. Permission from IEEE must be obtained for all other uses, in any current or future media, including reprinting/republishing this material for advertising or promotional purposes, creating new collective works, for resale or redistribution to servers or lists, or reuse of any copyrighted component of this work in other works.}

\maketitle
\thispagestyle{fancy}

\begin{abstract}
Transient stability control in smart grids requires rapid post-fault damping of generator frequency and rotor angle deviations to prevent cascading failures. This paper proposes FedPPO-PG, a Federated Multi-Agent Proximal Policy Optimization framework with Physics-Grounded neighborhoods, which reformulates transient stability control as a cooperative multi-agent reinforcement learning problem optimized directly against closed-loop stability objectives. Each generator hosts an independent local actor augmented with the frequency deviations of its two most strongly coupled electrical neighbors, identified from the post-fault Kron-reduced susceptance matrix. A guided policy initialization phase warm-starts all actors from the classical decentralized controller, while a centralized critic guides advantage estimation under the centralized training--decentralized execution (CTDE) paradigm. Evaluated on a simulation of the IEEE 39-bus benchmark system across five training and three unseen fault contingencies, FedPPO-PG achieves 100\% stabilization in all 24 trials, reduces mean stability time by \textbf{72.4\%}, and cuts the control power by 7-14 times compared to the centralized baseline. Each actor executes independently with no central coordinator at deployment, and the per-actor inference latency satisfies the IEEE/IEC 60255-118-1-2018 real-time reporting requirements.   
\end{abstract}


\section{Introduction}

Modern smart grids (SGs) are increasingly modeled as cyber-physical systems that integrate sensing, communication, and control to improve dependability and resilience in the face of disruptions. Transient stability control remains a crucial challenge; following a severe fault, generator rotor angles and frequencies must be stabilized rapidly to prevent cascading failures and loss of synchronism.

Recent work has explored interpretable neural architectures for distributed transient stability control. In \cite{2026TPECkan}, the centralized training–decentralized execution (CTDE) paradigm was employed by training spline-based Kolmogorov–Arnold Networks (KANs) and Chebyshev-based KANs (ChebyKANs) to approximate centralized control actions using only local PMU measurements. Closed-loop simulations revealed a significant robustness gap: low root-mean-squared error (RMSE) in offline evaluation did not consistently translate into improved dynamic resilience under unseen fault contingencies. Specifically, in fully decentralized deployment, neither neural variant could stabilize the system unless control authority was concentrated on high-inertia generators; in that case, ChebyKANs demonstrated better robustness.

Subsequently, a federated learning control (FLC) framework \cite{shahbaz2025flc} was proposed to enhance distributed coordination. In this approach, ChebyKAN-based local controllers collaboratively learned a shared global policy via federated averaging, improving generalization at moderate penetration levels. However, performance degraded at higher levels of distributed control penetration, so full distributed control was not achieved, and neural network inference latency remained a practical concern for real-time deployment.

A key finding from these prior studies is that closed-loop stability objectives are not directly optimized by supervised imitation of centralized control. Function approximation accuracy alone is insufficient to guarantee robust transient performance under unseen disturbances, because offline regression losses are not aligned with dynamic resilience metrics. This motivates reformulating transient stability control as a reinforcement learning (RL) problem, in which the policy is optimized directly against closed-loop performance criteria rather than a supervisory signal, and paves the way to surpass current control benchmarks and previous attempts to solve this problem.

In this work, we propose a federated multi-agent proximal policy optimization with Physics-Grounded Neighborhoods (FedPPO-PG) framework for transient stability control. Rather than imitating a centralized supervisor, each generator hosts an independent local actor trained through reward-driven policy gradients to directly optimize closed-loop stability and damping performance. Coordination is achieved through periodic federated averaging of actor weights, combined with centralized, critic-guided advantage estimation during training, while preserving the decentralized execution structure required for resilient deployment. The proposed framework is evaluated on multiple fault contingencies of the IEEE 39-bus benchmark. 

\section{System Model}
\label{sec:system_model}

\begin{figure*}[!ht]
\centering
\includegraphics[width=1.0\linewidth]{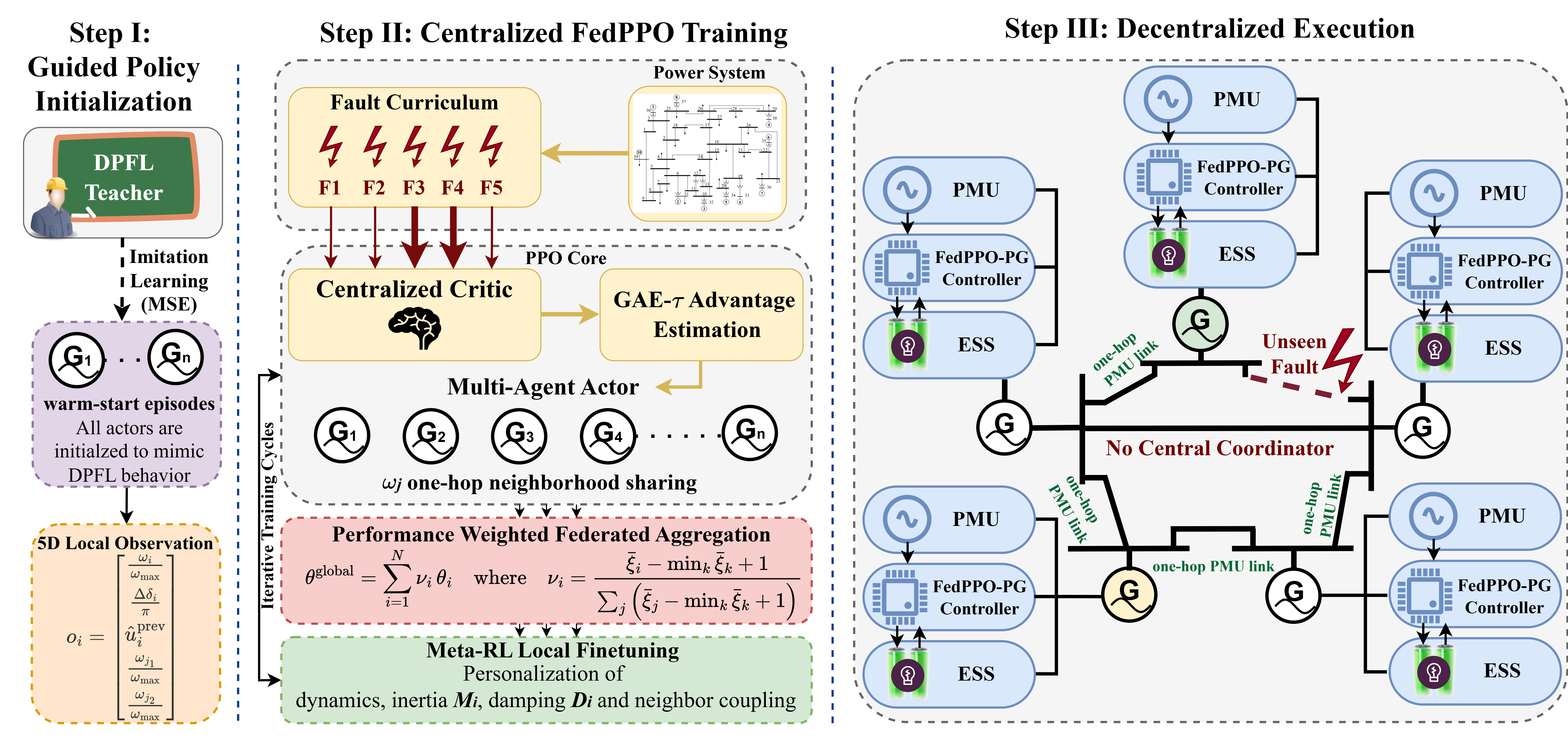}
\caption{The proposed FedPPO-PG framework: (I) behavior cloning warm-start from a decentralized PFL teacher, (II) centralized training with fault curriculum, performance-weighted federated aggregation, and Meta-RL local fine-tuning, and (III) fully decentralized execution using one-hop PMU communication with no central coordinator.}
\label{fig:MA_SG}
\end{figure*}

As shown in Fig.~\ref{fig:MA_SG}, the transient stability control problem is modeled as a multi-agent system over the SG transmission network. Each agent at a generator bus comprises a synchronous generator, a phasor measurement unit (PMU) for local state measurement, an Energy Storage System (ESS) with fast active-power charging and discharging capability, and a local controller that computes the ESS power command.

The electromechanical dynamics of interconnected synchronous generators are described by the classical swing equation~\cite{PS_Analysis}, which captures the time evolution of each generator's rotor angle \( \delta \) and angular frequency \( \omega \) driven by the balance between mechanical input power and electrical output power. For a power system with \(N\) generators, the dynamics of generator \(i\), \(\forall i \in \{1, \dots, N\}\), are given by

\begin{equation}\label{eqn: swing}
\dot{\delta}_i = \omega_i, \qquad
M_i \dot{\omega}_i = -D_i \omega_i + \bigl(P_{m,i} - P_{e,i} \bigr),
\end{equation}
where $\delta_i$ and $\omega_i$ denote the rotor angle and normalized angular frequency of generator $i$, with time derivatives $\dot{\delta}_i$ and $\dot{\omega}_i$, respectively, $M_i$ and $D_i$ are the inertia and damping coefficients. $P_{m,i}$ and $P_{e,i}$ denote the mechanical input and electrical output powers, respectively; the latter is expressed as:
\begin{equation}
\label{eqn: electrical power}
P_{e,i} = \sum_{k=1}^{N} |E_i|\,|E_k|
\left[ G_{ik} \cos(\delta_i - \delta_k) + B_{ik} \sin(\delta_i - \delta_k) \right],
\end{equation}
where \(|E_i|\) and \(|E_k|\) are the internal voltage magnitudes of generators \(i\) and \(k\), respectively, and \( G_{ik} \) and \( B_{ik} \) denote the equivalent Kron-reduced conductance and susceptance terms~\cite{kron_reduction}, representing the effective coupling between generators in the reduced network model.

This coupling structure is exploited in two ways in the proposed framework: the Kron-reduced susceptance magnitudes $|B_{ik}|$ identify each generator's most strongly coupled neighbors, defining the local observation vector, and the post-fault deviations in $\omega_i$ and $\delta_i$ quantify per-generator disturbance severity, guiding the federated aggregation weights, which are detailed in the following section.

As a baseline control scheme, we consider a centralized parametric feedback linearization (CPFL) controller~\cite{re_feedback_linearization}, which determines the auxiliary control input power at generator~$i$ bus as:
\begin{equation}\label{eqn: CPFL Pu}
P_{u,i} = -\left( P_{a,i} - P_{d,i} \right),
\end{equation}
where $P_{a,i} = P_{m,i} - P_{e,i}$ denotes the accelerating power. The decentralized PFL (DPFL) component utilizes only local PMU measurements and is given by
\begin{equation}\label{eqn: DPFL Pu}
P_{d,i} = -\left( \alpha_i \omega_i + \beta_i (\delta_i - \delta_i^*) \right),
\end{equation}
with $\alpha_i, \beta_i \geq 0$ representing frequency and angle stabilization gains, and $\delta_i^*$ denoting the desired rotor angle setpoint.

A negative \(P_{u,i}\) corresponds to absorbing active power from the bus (charging the ESS), while a positive value corresponds to injecting power into the bus (discharging the ESS).

\section{Methodology}

\subsection{Reinforcement Learning Problem Formulation}

The transient stability control task is cast as a cooperative multi-agent Markov decision process (MAMDP)~\cite{mamdp_ref} over $N{=}10$ generator agents.  At each discrete time step~$k$, agent~$i$ observes a local vector $o_{i,k}$, selects a continuous control action~$a_{i,k}$, and the environment transitions according to the RK2-discretized swing equations of Section~\ref{sec:system_model}. A shared scalar reward, $ r_k$, is returned to all agents.  The objective is a set of decentralized policies $\{\pi_{\theta_i}\}_{i=1}^{N}$ that maximise the expected discounted return $\mathbb{E}[\sum_{k=0}^{K}\gamma^k r_k]$, where $\gamma$ is the discount factor.

\subsection{Observation Space}

\subsubsection{Physics-Grounded neighborhood Selection}

A simple decentralized design would restrict each agent to purely local measurements $(\omega_i, \delta_i)$ \cite{shahbaz2025flc, 2026TPECkan}.  We argue, however, that such a restriction disregards the network's electromechanical coupling structure.  The electrical power exchanged between generators $i$ and $k$ is proportional to the Kron-reduced susceptance $|B_{ik}|$ (see equation~\ref{eqn: electrical power}); a fault that perturbs generator~$k$ transmits accelerating power to generator~$i$ in direct proportion to $|B_{ik}|$.  Ignoring the most strongly coupled neighbors, therefore, discards the dominant physical signal available to the agent at the moment it most needs
to act.

Therefore, we introduce \textbf{physics-grounded neighborhood selection} to determine each agent's communication partners directly from the power-flow equations. The top-$K$ neighbors of generator~$i$ are selected from the post-fault Kron-reduced susceptance matrix $\mathbf{B}_{\mathrm{post}}$ as
\begin{equation}\label{eqn:nbr}
  \mathcal{N}_i =
    \operatorname*{arg\,top\text{-}K}_{j \neq i}\;|B_{ij}|,
\end{equation}
where $K$ is a design parameter; we set $K{=}2$ for simplicity throughout this work. The normalized frequency deviations of the selected neighbors, $\omega_j/\omega_{\max}$ for $j \in \mathcal{N}_i$, are appended to the local observation. Because the susceptances are derived from the actual reduced admittance matrix, the neighborhood is \emph{fault-aware}: it is recomputed for every fault contingency, so the agent always receives the frequencies of the generators whose dynamics are most tightly coupled to its own under the prevailing post-fault topology.  The resulting observation remains local in the operational sense; it requires only one-hop direct PMU-to-PMU communication, but it is \emph{physically consistent} with the underlying network physics in a way that pure single-bus observation is not.  This design corresponds to $k{=}1$ hop decentralized execution in the MARL literature~\cite{k_hop_ref}, and the execution-time claim of this paper is that each agent acts using only its own PMU measurements and the one-hop neighbor frequencies obtained via direct PMU sharing, with no central coordinator. 

\subsubsection{Local Observation Vector}
The normalized five-dimensional observation of agent~$i$ is
\begin{equation}\label{eqn:obs}
  o_i = \Bigl[
    \tfrac{\omega_i}{\omega_{\max}},\;
    \tfrac{\Delta\delta_i}{\pi},\;
    \hat{u}_i^{\mathrm{prev}},\;
    \tfrac{\omega_{j_1}}{\omega_{\max}},\;
    \tfrac{\omega_{j_2}}{\omega_{\max}}
  \Bigr]^{\!\top} \in \mathbb{R}^{2K+3},
\end{equation}
where $\{j_1, \dots, j_K\} = \mathcal{N}_i$ are the physics-grounded neighbors from \eqref{eqn:nbr}, $\omega_{\max}$ is a normalization constant, and $\hat{u}_i^{\mathrm{prev}}\!\in[-1,1]$ is the normalized ESS command from the previous time step.

The angle deviation $\Delta\delta_i$ is referenced to a post-fault SEP estimate $\bar{\delta}_i$ once the system has sufficiently settled following fault clearance, and to the pre-fault equilibrium $\delta_i^*$ otherwise:
\begin{equation}\label{eqn:sep}
  \Delta\delta_i =
    \delta_i - \begin{cases}
      \bar{\delta}_i & \text{post-fault (settled)},\\
      \delta_i^*     & \text{otherwise.}
    \end{cases}
\end{equation}
This adaptive reference is particularly important for fault scenarios in which post-fault topologies shift the reachable SEP significantly away from the pre-fault equilibrium.

\subsubsection{Global State (Training Only)}

During training, the centralized critic receives
\begin{equation}
  s = [o_1^\top,\, o_2^\top,\, \ldots,\, o_{N}^\top]^\top
    \in \mathbb{R}^{50}.
\end{equation}
This vector is \emph{never} available to any actor at execution time.

\subsection{Action Space}

Each agent outputs a scalar $a_i \in [-1, 1]$, mapped to a physical
ESS power command as
\begin{equation}\label{eqn:action}
  P_{u,i} = a_i \cdot u_{\max,i}, \qquad
  u_{\max,i} = \min\bigl(\alpha\,|P_{m,i}|,\; P_{\max}\bigr)\;\mathrm{pu},
\end{equation}
where $\alpha$ is a fraction of the generator's mechanical power output and $P_{\max}$ is a hard ceiling. This ensures that ESS commands remain within the physical operating limits of each generator, with the \texttt{tanh} output layer of the actor network guaranteeing $|a_i| \leq 1$ by construction.

\subsection{Reward Function}
A shaped scalar reward is shared by all agents at every step $k$:
\begin{align}\label{eqn:reward}
  r_k = &-\lambda_\omega \sum_{i=1}^{N}\!\hat{\omega}_i^2
         - \lambda_\delta \sum_{i=1}^{N}\!\widehat{\Delta\delta}_i^2
         - \lambda_u      \sum_{i=1}^{N}\!\hat{u}_i^2  \notag\\
        &- \lambda_{du}   \sum_{i=1}^{N}\!
           \bigl(\hat{u}_i - \hat{u}_i^{\mathrm{prev}}\bigr)^2
         + r_{\mathrm{term}},
\end{align}
with $\hat{\omega}_i{=}\omega_i/\omega_{\max}$, $\widehat{\Delta\delta}_i{=}\Delta\delta_i/\pi$, $\hat{u}_i{=}P_{u,i}/u_{\max,i}$, and penalty weights $\lambda_\omega \gg \lambda_\delta \gg \lambda_u, \lambda_{du} > 0$ tuned so that frequency stabilization dominates the reward signal. The control-rate penalty $\lambda_{du}$ suppresses high-frequency chattering in the ESS commands. The terminal component is
\begin{equation}
  r_{\mathrm{term}} = \begin{cases}
    +R_s + c\,(T - t_k) & \text{stabilized,}\\
    -R_f                 & \text{loss-of-synchronism,}\\
    0                    & \text{timeout,}
  \end{cases}
\end{equation}
where $R_s$ and $R_f$ are the stabilization bonus and failure penalty, respectively, and $c$ is a time-bonus coefficient that rewards faster convergence. Stabilization is declared when $|\hat{\omega}_i| < \epsilon_\omega$ for all $i$ simultaneously over $H$ consecutive steps, where $\epsilon_\omega$ and $H$ are design parameters. Reward weights are selected to reflect the relative importance of frequency regulation, angle deviation, and control effort in transient stability applications.

Per-agent disturbance contributions 
\begin{equation}
    \xi_{i,k} = -(\lambda_\omega\hat{\omega}_i^2 + 
    \lambda_\delta\widehat{\Delta\delta}_i^2)
\end{equation}
are computed at each step to support the performance-weighted federated aggregation described in Section~\ref{sec:fedppo}. Intuitively, $\xi_{i,k}$ reflects how severely generator $i$ is disturbed at step $k$: agents with larger frequency and angle deviations accumulate higher aggregation weights, biasing the global model update toward those that need it most.

\subsection{Proposed FedPPO Algorithm}
\label{sec:fedppo}
The proposed algorithm, \emph{Imitation-Bootstrapped Federated PPO with Physics-Grounded neighborhoods} (FedPPO-PG), consists of three interleaved phases, as shown in Algorithm~\ref{alg:fedppo} and detailed below.

\subsubsection{Guided Policy Initialization}

Training a multi-agent PPO policy from random initialization in a safety-critical physical environment is difficult~\cite{schulman2017proximalpolicyoptimizationalgorithms}. Random actors rarely stabilize the system, yielding a degenerate reward landscape dominated by the failure penalty. To avoid this, we warm-start all $N$ actors via guided policy initialization from the classical decentralized DPFL controller (Eq.~\eqref{eqn: DPFL Pu}).

The teacher produces a normalized target action 
$a_i^{\mathrm{DPFL}} = \operatorname{clip}(P_{u,i}^{\mathrm{DPFL}}/ 
u_{\max,i}, -1, 1)$, and each actor is updated by minimizing
\begin{equation}\label{eqn:gp}
  \mathcal{L}_{\mathrm{GP}}^{(i)} =
    \bigl\|\tanh\bigl(\mu_{\theta_i}(o_i)\bigr)
           - a_i^{\mathrm{DPFL}}\bigr\|^2.
\end{equation}
MSE on the mean output is used rather than negative log-likelihood so that the actor's log-standard deviation is not collapsed during initialization, preserving exploration capacity for subsequent RL. After the warm-start phase, every actor begins from a region of parameter space where the DPFL stabilization heuristic is already encoded, providing a non-degenerate reward signal from the very first RL episode.

\subsubsection{Federated PPO}

The main training loop is an on-policy PPO procedure operating under the CTDE paradigm.

\paragraph{Network architecture}
The centralized critic $V_\phi(s)$ is a multi-layer perceptron (MLP) mapping the $5N$-dimensional global state to a scalar value estimate. Each local actor $\pi_{\theta_i}$ is an independent MLP mapping the 5-dimensional observation $o_i$ to a Gaussian distribution over the action space, with \texttt{tanh} squashing to enforce the $[-1,1]$ action bound.  All weight matrices are initialized orthogonally to promote stable early-training gradient flow.

\paragraph{Advantage estimation}
At each step the critic evaluates $V_\phi(s_k)$; advantages are computed via Generalized Advantage Estimation (GAE)-$\tau$~\cite{gae_ref}. A key implementation detail is that the bootstrap value is set to zero only on \emph{true terminal} transitions (loss-of-synchronism or stabilization) and to $V_\phi(s_{K})$ on timeouts, preventing systematic underestimation of returns at episode boundaries.

\paragraph{Per-actor PPO update}
Each actor is updated by the clipped surrogate objective:
\begin{equation}\label{eqn:ppo}
  \mathcal{L}_{\pi}^{(i)} = -\mathbb{E}_t\!\Bigl[
    \min\bigl(\rho_t^{(i)}\,\tilde{A}_t^{(i)},\;
    \mathrm{clip}(\rho_t^{(i)},\,1{\pm}\epsilon)\,\tilde{A}_t^{(i)}\bigr)
    - \beta_t\,\mathcal{H}[\pi_{\theta_i}]
  \Bigr],
\end{equation}
where $\rho_t^{(i)} = \pi_{\theta_i}(a_t^{(i)}|o_t^{(i)}) /
\pi_{\theta_i^{\mathrm{old}}}(a_t^{(i)}|o_t^{(i)})$ is the likelihood ratio, $\epsilon{=}0.2$, and
$\beta_t$ is an entropy coefficient that is annealed during training to balance exploration early on with policy sharpening at convergence.  The per-actor advantage $\tilde{A}_t^{(i)} = \hat{A}_t \cdot w_t^{(i)} \cdot N$ uses a
physics-grounded weighting
\begin{equation}\label{eqn:agentweight}
  w_t^{(i)} = \frac{|\xi_{i,t}|}{\sum_j |\xi_{j,t}|},
\end{equation}
so that agents exhibiting larger frequency deviations, and therefore contributing more to the instability, receive proportionally stronger gradient signals.

\paragraph{Learning Curriculum}

Training episodes sample all five fault contingencies from  Table~\ref{tab:faults}, with F3 and F4 oversampled, and fault-clearing times randomized over a predefined range.

\subsubsection{Performance-Weighted Federated Averaging}

Every $T_{\mathrm{fed}}$ PPO episodes, the actor weight tensors are aggregated into a single global model via weighted FedAvg:
\begin{equation}\label{eqn:fedavg}
 \theta^{\mathrm{global}} = \sum_{i=1}^{N} \nu_i\,\theta_i, \qquad
  \nu_i = \frac{\bar{\xi}_i - \min_k\bar{\xi}_k + 1}
               {\sum_j(\bar{\xi}_j - \min_k\bar{\xi}_k + 1)},
\end{equation}
where $\bar{\xi}_i$ is actor~$i$'s mean per-agent contribution over the most recent k-episodes window. The shift-by-minimum with additive unity maps all weights to the positive real line before normalization. Actors that have specialized successfully, evidenced by higher (less negative) cumulative rewards, contribute proportionally more to the shared global prior.  The global model is then redistributed identically to all 10 actors.

\subsubsection{Meta-RL-Inspired Local Fine-Tuning}

Immediately after redistribution, each actor undergoes $T_{\mathrm{tune}}$ additional local PPO episodes without a subsequent averaging step. This inner loop is directly inspired by model-agnostic meta-learning (MAML)~\cite{maml_ref} and personalized federated learning~\cite{per_fed_ref}: $\theta^{\mathrm{global}}$ serves as a shared prior that lies in a region of parameter space from which rapid per-generator adaptation is possible.  Because generator inertias typically span different orders of magnitude
, a uniform global model is insufficient for optimal control; therefore, the fine-tuning windows allow each actor to re-specialize to its generator's unique inertia, damping, and physics-grounded coupling profile before the next federated round.

The cycle repeats until convergence, with the number of FedAvg rounds and local fine-tuning episodes per actor determined by $T_{\mathrm{fed}}$ and $T_{\mathrm{tune}}$, respectively.  This combination of guided policy initialization from the DPFL controller, physics-grounded neighborhood selection via post-fault susceptance coupling, cooperative reward shaping with per-agent counterfactual advantage weighting, performance-weighted federated averaging, and meta-RL-inspired local re-specialization constitutes the core innovation of this work. The complete procedure is given in Algorithm~\ref{alg:fedppo} and illustrated in Figure~\ref{fig:MA_SG}.





\begin{algorithm}[t]
\caption{FedPPO-PG: Imitation-Bootstrapped Federated PPO with Physics-Grounded neighborhoods}
\label{alg:fedppo}
\begin{algorithmic}[1]
\Statex \textbf{--- Initialization ---}
\State $\mathcal{N}_i \leftarrow \operatorname{top-}K_{j \neq i}|B_{ij}|$ 
       \textbf{for all} $i$ 
       \State \Comment{physics-grounded neighbors}
\State Initialize $\{\pi_{\theta_i}\}_{i=1}^{N}$, $V_\phi$
\Statex \textbf{--- Guided Policy Initialization ---}
\For{$e = 1, \dots, T_{\mathrm{GP}}$}
    \State $a_i^{\mathrm{DPFL}} \leftarrow 
           \operatorname{clip}(P_{u,i}^{\mathrm{DPFL}}/u_{\max,i},\,-1,\,1)$
    \State \Comment{teacher action}
    \State $\theta_i \leftarrow \theta_i - 
           \nabla_{\theta_i}\mathcal{L}_{\mathrm{GP}}^{(i)}$ 
           \textbf{for all} $i$ 
           \State \Comment{behavior cloning via MSE}
\EndFor
\Statex \textbf{--- FedPPO-PG Loop ---}
\For{$e = 1, \dots, T_{\mathrm{total}}$}
    \State Sample fault $f \sim p_{\mathrm{curriculum}}$,\; 
           $t_{cf} \sim \mathcal{U}[t_{\min}, t_{\max}]$
    \State Collect rollout: $a_i \sim \pi_{\theta_i}(\cdot \mid o_i)$,\;
           $\hat{V} \leftarrow V_\phi(s)$
    \State Compute $\hat{A}_t$ via GAE-$\tau$;\;
           $\tilde{A}_t^{(i)} \leftarrow \hat{A}_t \cdot \nu_i^{(t)} \cdot N$
    \State \Comment{per-agent advantage weighting}
    \State Update $V_\phi$ via MSE on returns
    \State Update each $\pi_{\theta_i}$ via 
           $\mathcal{L}_{\pi}^{(i)}$ \eqref{eqn:ppo}
    \If{$e \bmod T_{\mathrm{fed}} = 0$}
        \State $\nu_i \leftarrow 
               (\bar{\xi}_i - \min_k \bar{\xi}_k + 1) \;/\; 
               \textstyle\sum_j(\bar{\xi}_j - \min_k\bar{\xi}_k+1)$
        \State \Comment{performance-weighted aggregation}
        \State $\theta^* \leftarrow \sum_{i=1}^{N} \nu_i\,\theta_i$
        \State $\theta_i \leftarrow \theta^*$ \textbf{for all} $i$
        \For{$t = 1,\dots,T_{\mathrm{tune}}$}
            \State Collect rollout and update $\pi_{\theta_i}$ 
                   \textbf{for all} $i$ 
            \State \Comment{Meta-RL local fine-tuning}
        \EndFor
    \EndIf
\EndFor
\State \Return $\{\pi_{\theta_i}\}_{i=1}^{N}$
       \hfill\Comment{$V_\phi$ discarded at deployment}
\end{algorithmic}
\end{algorithm}
\vspace{-3mm}
\section{Results and Discussion}

\subsection{Simulation Setup}

This work is evaluated on the IEEE 39-bus New England test system~\cite{39bus_ref}, which consists of 10 synchronous generators and 39 buses, and serves as the benchmark for all simulated experiments. Swing equation dynamics are discretized via the second-order Runge-Kutta (RK2) method with $\Delta t{=}0.01$\,seconds (s) over a 100\,seconds simulation window. 

A three-phase bus fault from Table~\ref{tab:faults} is applied at $t{=}0.5$\,s and cleared by tripping the associated line at $T_{cf} = 0.5 + t_{cf}$\,, where $t_{cf} \in \{0.15, 0.20, 0.25\}$\,s spans moderate to severe clearing delays.
\begin{wraptable}[12]{r}{0.28\textwidth}
\vspace{0pt}
\centering
\small{
\setlength{\tabcolsep}{2pt}
\renewcommand{\arraystretch}{0.8}
\caption{Fault Details}
\label{tab:faults}
\begin{tabular}{@{}ccc@{}}
\toprule
\textbf{Fault} & \textbf{Faulted Bus} & \textbf{Faulted Line} \\
\midrule
F1 &  17 & 17--18 \\
F2 &  11 & 10--11 \\
F3 &  22 & 21--22 \\
F4 &  29 & 28--29 \\
F5 &   5 &  5--8  \\
\midrule
F6$^{\dagger}$ & 16 & 16--17 \\
F7$^{\dagger}$ & 26 & 26--27 \\
F8$^{\dagger}$ &  2 &  2--3  \\
\bottomrule
\multicolumn{3}{@{}l}{\tiny $^{\dagger}$Unseen.}
\end{tabular}
\vspace{6pt}
}
\end{wraptable}
The ESS controller activates at fault clearance ($T_{sc} = T_{cf}$); no control is applied during the fault. A trial is considered stabilized if $|\omega_i| < \epsilon_\omega$ for all generators simultaneously over $H$ consecutive steps and the stability time is recorded accordingly; otherwise the controller is reported as unstabilized. The FedPPO-PG hyperparameters used in all reported experiments are summarized in Table~\ref{tab:hyperparams}.

\begin{table}[t]
\vspace{1mm}
\caption{FedPPO-PG Training Hyperparameters}
\label{tab:hyperparams}
\centering
\footnotesize
\setlength{\tabcolsep}{4pt}
\renewcommand{\arraystretch}{1.1}

\begin{tabular}{p{0.65\linewidth}c}
\toprule
\textbf{Parameter} & \textbf{Value} \\
\midrule
Optimizer & Adam ($\epsilon{=}10^{-5}$) \\
Learning rates (actor/critic) & $10^{-4} / 10^{-3}$ \\
Gradient clip norm & 0.5 \\
PPO epochs / mini-batches & 4 / 4 \\
Discount $\gamma$, GAE $\tau$ & 0.99 / 0.95 \\
PPO clip $\epsilon$ & 0.2 \\
Hidden layers (actor/critic) & $2\times128 / 2\times256$ \\
Entropy coef.\ $\beta$ (init/final) & 0.01 / 0.001 \\
Reward weights $(\lambda_\omega,\lambda_\delta,\lambda_u,\lambda_{du})$ & 20 / 5 / 0.01 / 0.005 \\
$R_s$, $R_f$, $c$ & 200 / 500 / 2.0 \\
$\epsilon_\omega$, $H$ & 0.01\,pu / 20 \\
Reward scale & 100 \\
Neighborhood size $K$ & 2 \\
Action bounds $(\alpha, P_{\max})$ & 0.25 / 0.5\,pu \\
$\omega_{\max}$ & 0.20\,pu \\
FedAvg $T_{\mathrm{fed}}$, Tune $T_{\mathrm{tune}}$ & 100 / 20 \\
GP warm-start / Episodes & 100 / 5000 \\
Fault weights (F1--F5) & 1.0, 1.0, 2.5, 2.5, 1.0 \\
Fault clearing time & $[0.10,\,0.40]$\,s \\
\bottomrule
\end{tabular}
\end{table}

\subsection{Closed-Loop Stability Performance}

Table~\ref{tab:stability} reports the mean stability time averaged across all three clearing offsets for each fault and controller. The DPFL controller fails to stabilize in every trial (0\% stabilization rate), confirming that its fixed linear structure provides insufficient damping under three-phase fault contingencies with line tripping. The  CPFL controller achieves 100\%
stabilization but at a mean stability time of 43.24\,s, requiring full system state and accelerating power from a central coordinator.
\begin{table}[h]
\renewcommand{\arraystretch}{1.3}
\caption{Mean Stability Time (s) Averaged Over $t_{\mathrm{cf}} \in \{0.15, 0.20, 0.25\}$\,s. ``--'' denotes failure to stabilize.}
\label{tab:stability}
\centering
\setlength{\tabcolsep}{4pt}
\begin{adjustbox}{max width=\columnwidth}
\begin{tabular}{lcccccccccc}
\toprule
\textbf{Controller} &
\textbf{F1} & \textbf{F2} & \textbf{F3} & \textbf{F4} & \textbf{F5} &
\textbf{F6}$^{\dagger}$ & \textbf{F7}$^{\dagger}$ & \textbf{F8}$^{\dagger}$ &
\textbf{Overall} \\
\midrule
CPFL     & 29.87 & 38.89 & 48.76 & 65.58 & 39.69 & 44.98 & 57.86 & 20.27 & 43.24 \\
DPFL     & --    & --    & --    & --    & --    & --    & --    & --    & --    \\
FedPPO-PG   & \textbf{3.01} & \textbf{11.41} & \textbf{30.95} & \textbf{20.50} &
           \textbf{1.12} & \textbf{9.94}  & \textbf{7.05}  & \textbf{11.58} &
           \textbf{11.95} \\
\bottomrule
\multicolumn{10}{l}{$^{\dagger}$Unseen faults: excluded from training distribution.}
\end{tabular}
\end{adjustbox}
\end{table}

FedPPO-PG achieves a 100\% stabilization rate across all 24 trials, including the three unseen fault scenarios F6--F8, reducing the mean stability time to \textbf{11.95\,s}, a \textbf{72.4\%} reduction relative to the CPFL baseline, while operating with only one-hop PMU communication and no central coordinator.

On the simpler contingencies F1 and F5, FedPPO-PG stabilizes in under 3\,s and 1.2\,s respectively. The hardest training fault, F3 (Bus~22, line~21--22 trip), takes 30.95\,s on average, reflecting the significant post-fault topology change that shifts the synchronous equilibrium. Generalization to unseen faults is strong: F6--F8 stability times (7.05--11.58\,s) are comparable to or better than the seen training faults F2 and F4, demonstrating that physics-grounded neighborhood selection and federated averaging together produce policies that transfer across fault locations.

\begin{figure}[t]
\centering
\includegraphics[width=0.95\linewidth]{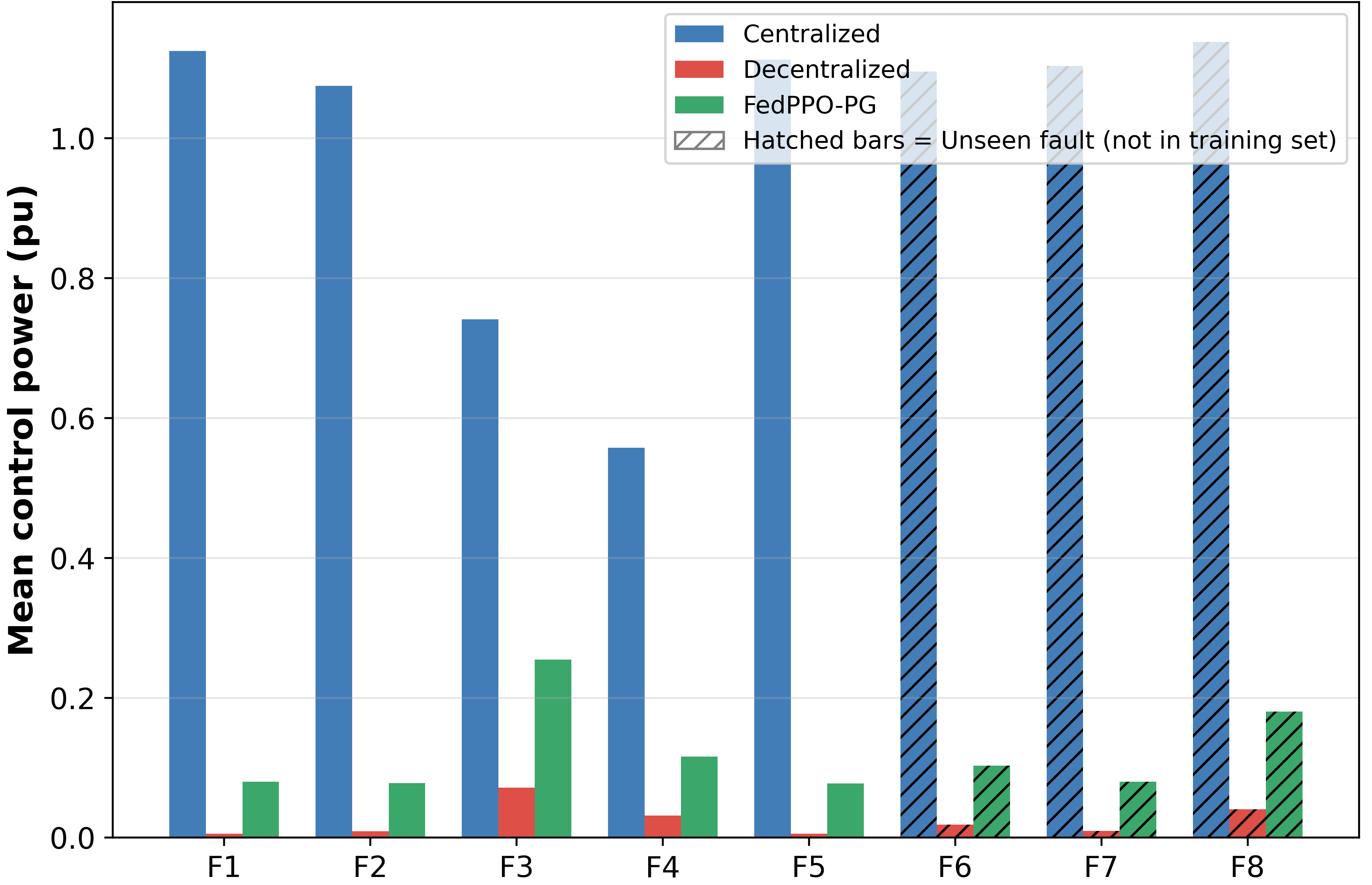}
\caption{Mean absolute control power for all controllers across F1--F8 at $t_{\mathrm{cf}}{=}0.2$\,s. DPFL fails to stabilize in all cases (labeled ``not stab''). Hatched bars represent unseen faults.}
\label{fig:effort_bar}
\end{figure}

Fig.~\ref{fig:effort_bar} compares the mean absolute control power across all scenarios at $t_{\mathrm{cf}}{=}0.2$\,s. CPFL demands 0.56--1.14\,pu across F1--F8, while FedPPO-PG achieves stabilization at 0.08--0.25\,pu (a 7--14$\times$ reduction) reflecting the learned policy's ability to apply targeted, distributed damping rather than the sustained large-magnitude of power charging/discharging characteristic of the baseline centralized controller.

\begin{figure*}[t]
\centering
    \begin{subfigure}{0.32\textwidth}
        \centering
        \includegraphics[width=\linewidth]{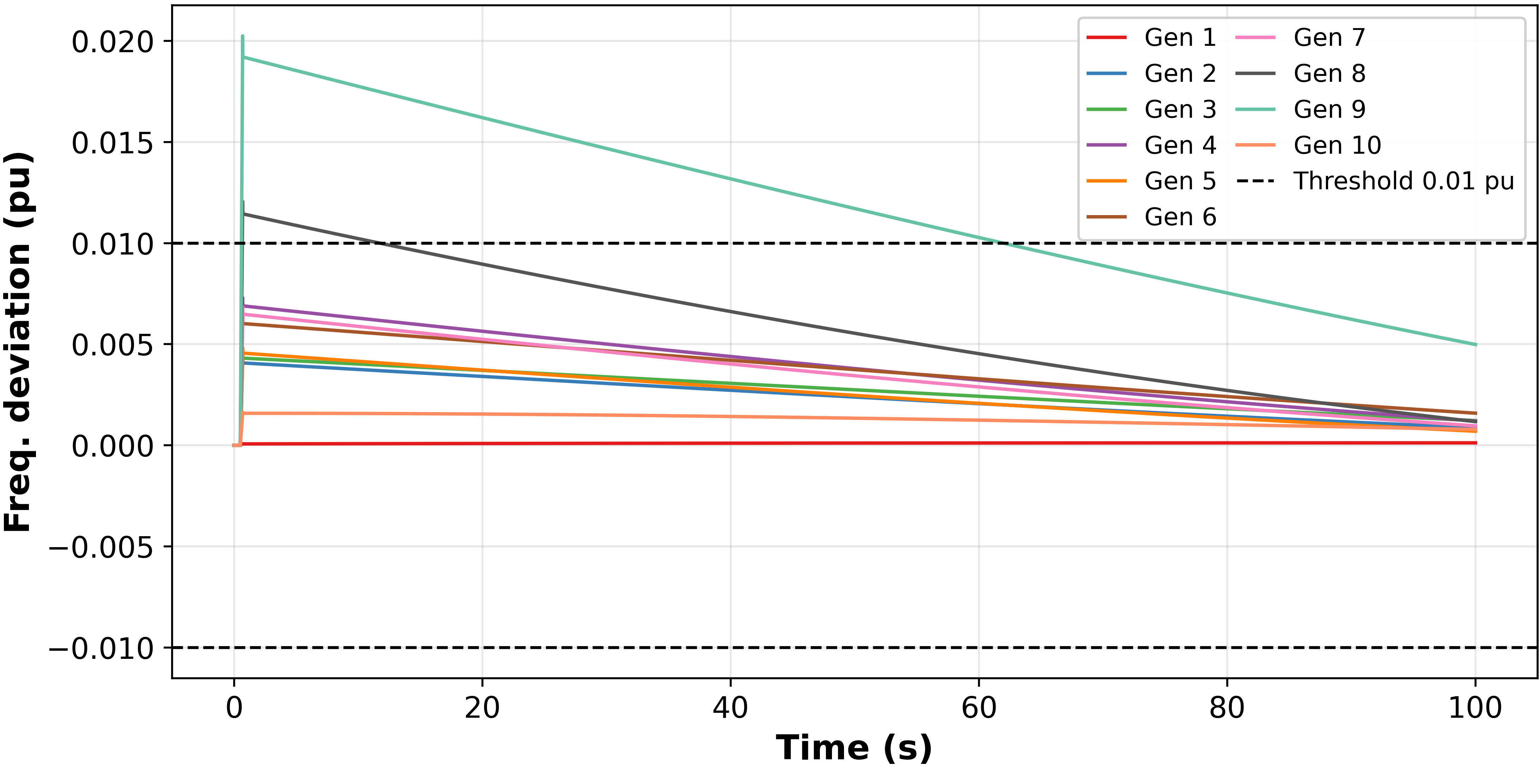}
        \caption{CPFL: frequency deviation.}
        \label{fig:f7_omega_cpfl}
    \end{subfigure}\hfill
    \begin{subfigure}{0.32\textwidth}
        \centering
        \includegraphics[width=\linewidth]{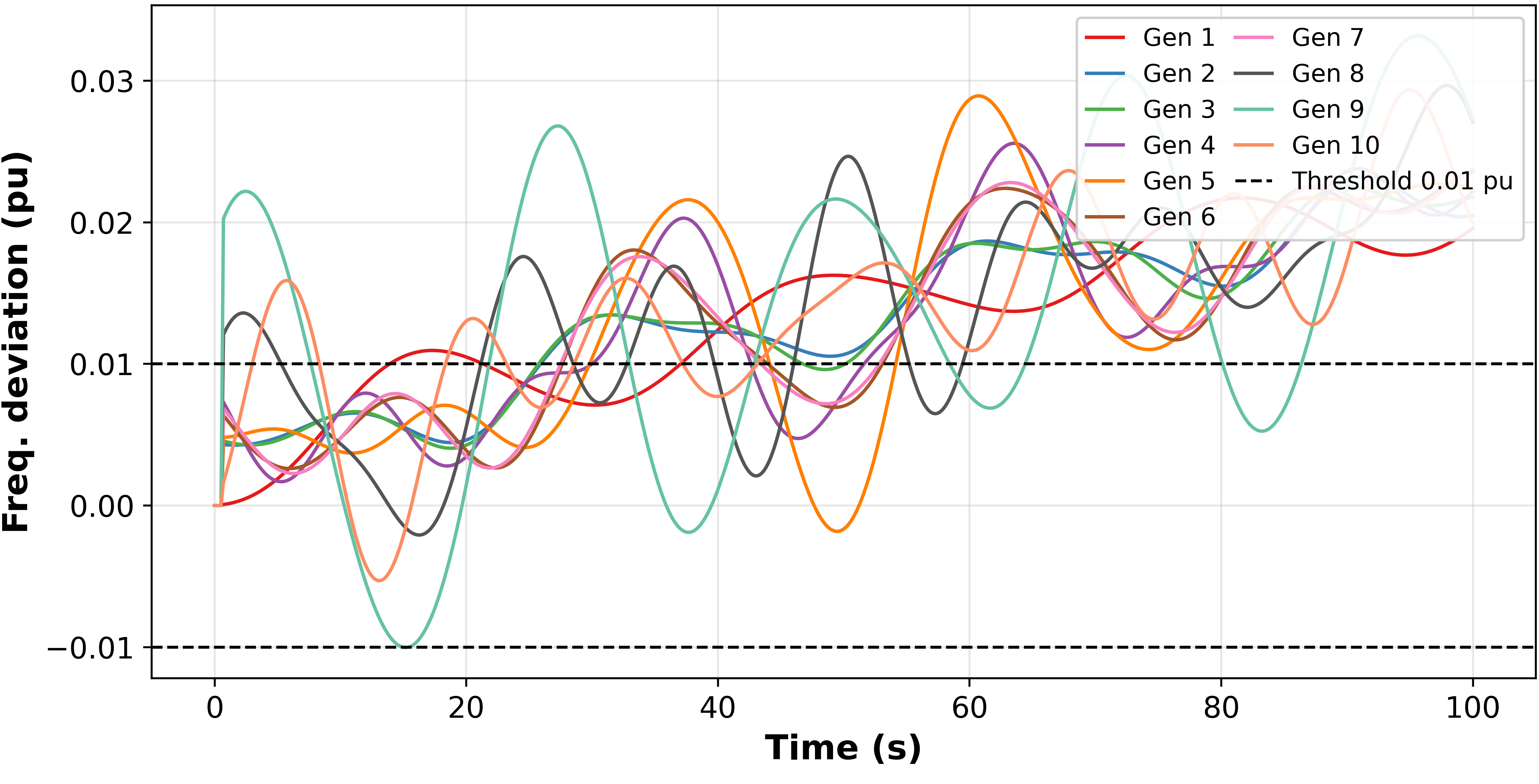}
        \caption{DPFL: frequency deviation.}
        \label{fig:f7_omega_dpfl}
    \end{subfigure}\hfill
    \begin{subfigure}{0.32\textwidth}
        \centering
        \includegraphics[width=\linewidth]{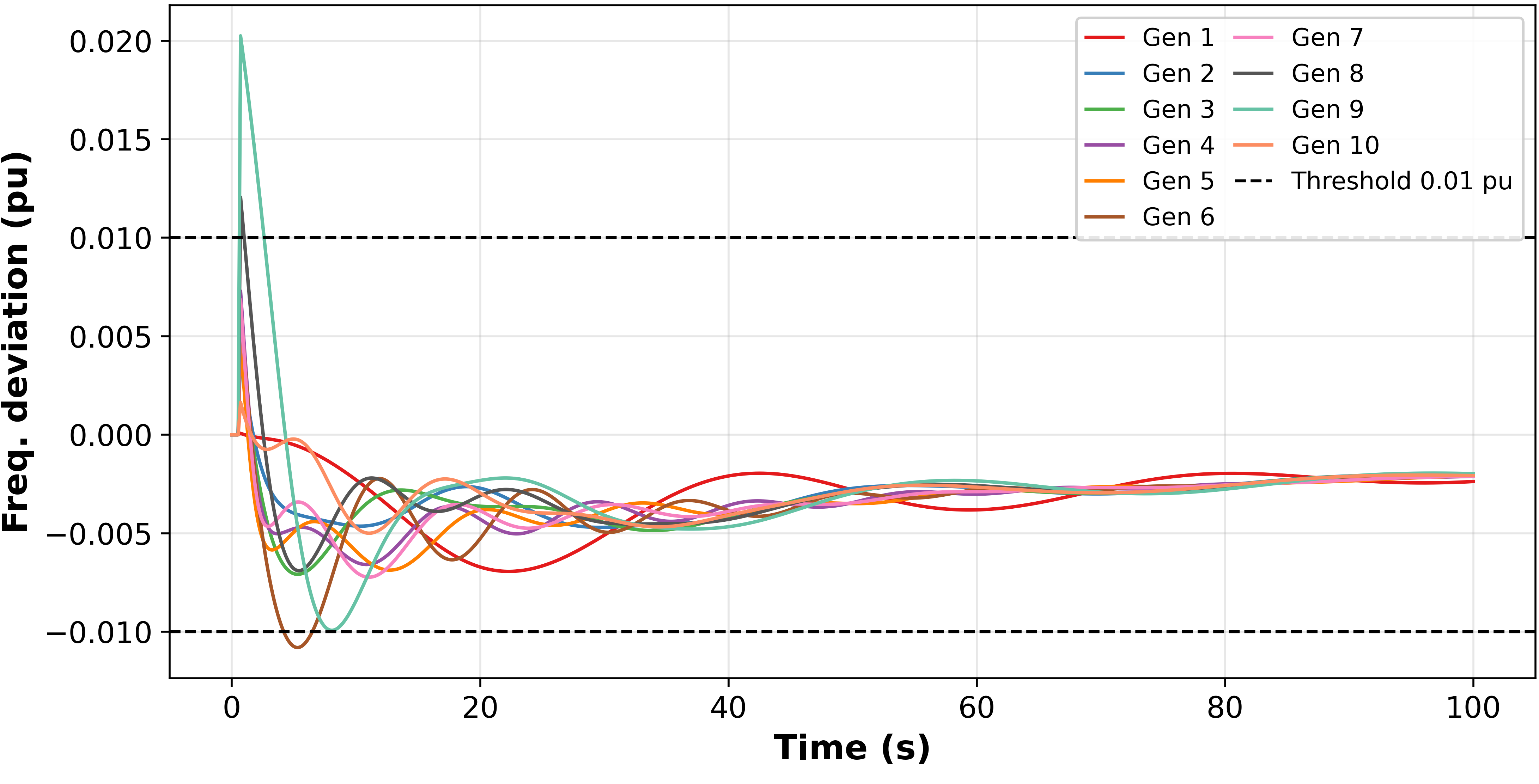}
        \caption{FedPPO-PG: frequency deviation.}
        \label{fig:f7_omega_fedppo}
    \end{subfigure}

    \vspace{2pt}

    \begin{subfigure}{0.32\textwidth}
        \centering
        \includegraphics[width=\linewidth]{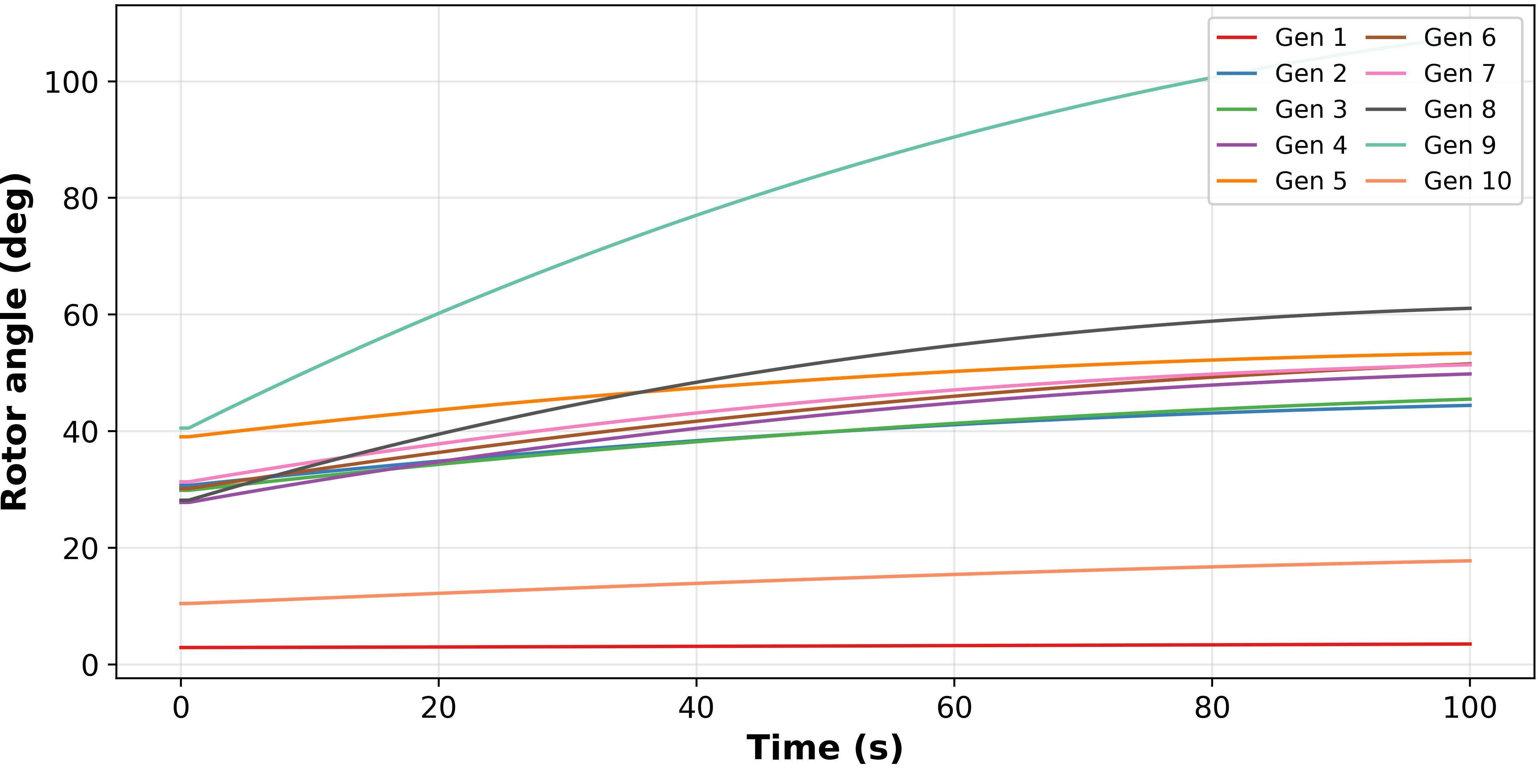}
        \caption{CPFL: rotor angle.}
        \label{fig:f7_phase_cpfl}
    \end{subfigure}\hfill
    \begin{subfigure}{0.32\textwidth}
        \centering
        \includegraphics[width=\linewidth]{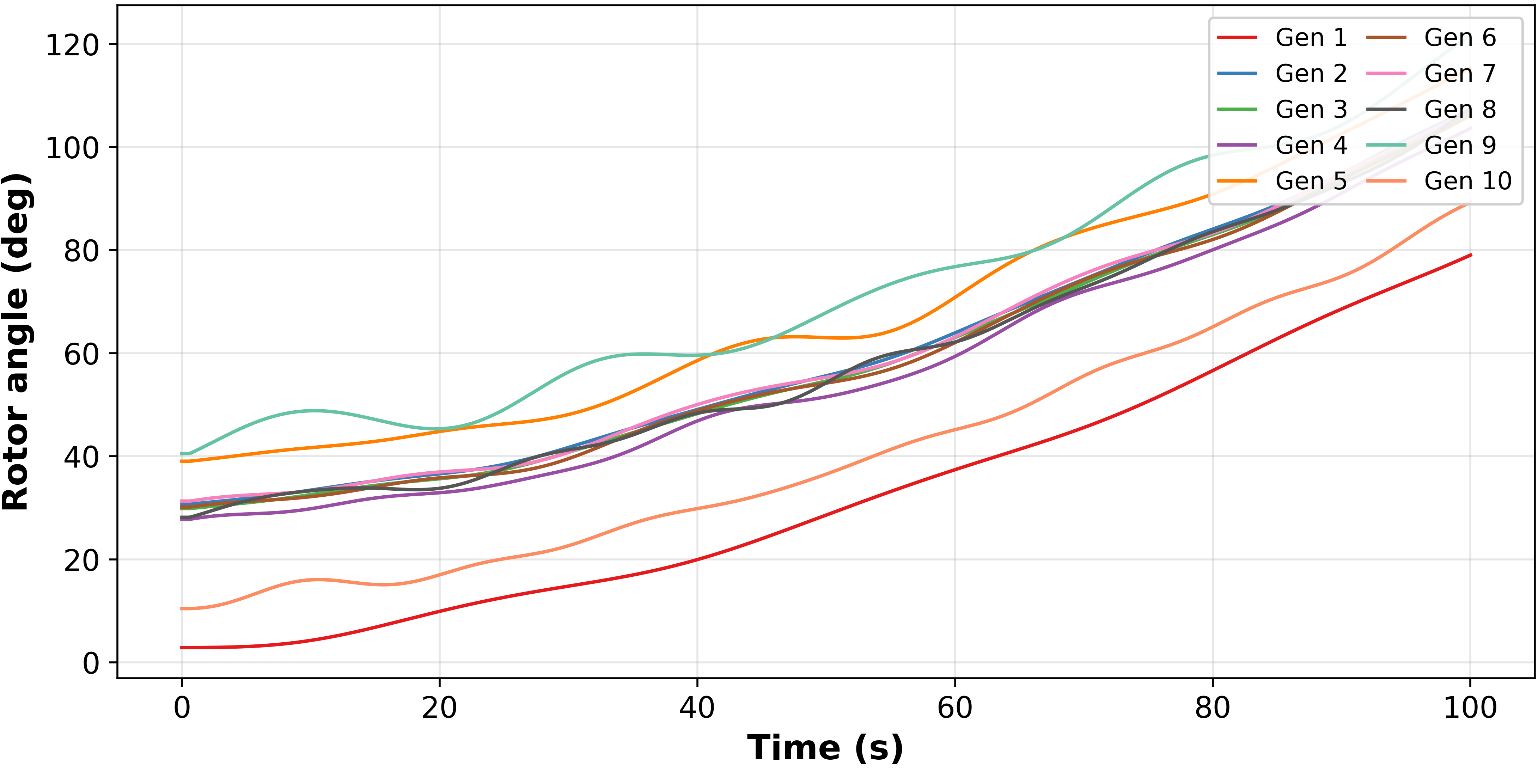}
        \caption{DPFL: rotor angle.}
        \label{fig:f7_phase_dpfl}
    \end{subfigure}\hfill
    \begin{subfigure}{0.32\textwidth}
        \centering
        \includegraphics[width=\linewidth]{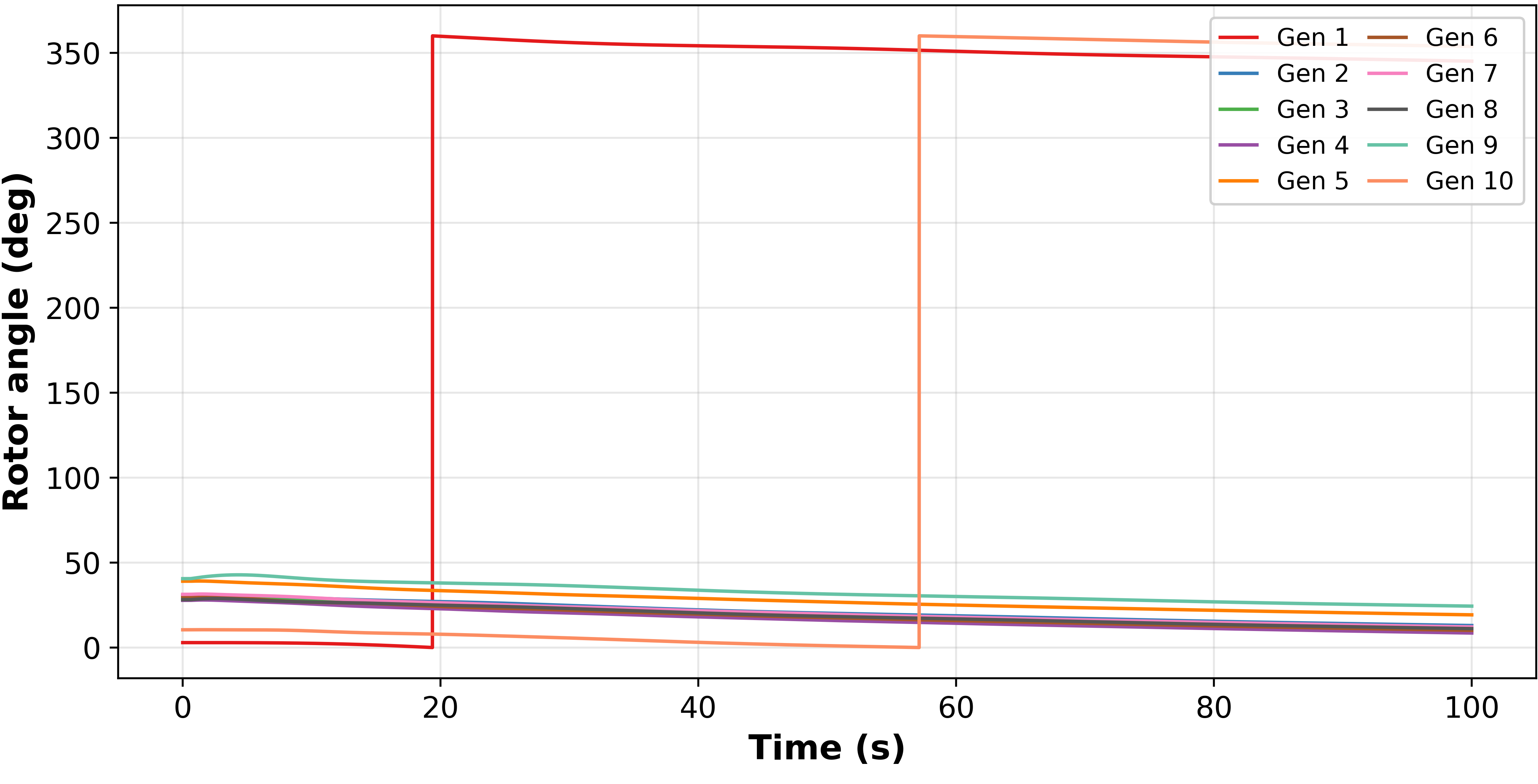}
        \caption{FedPPO-PG: rotor angle.}
        \label{fig:f7_phase_fedppo}
    \end{subfigure}

    \vspace{2pt}

    \begin{subfigure}{0.32\textwidth}
        \centering
        \includegraphics[width=\linewidth]{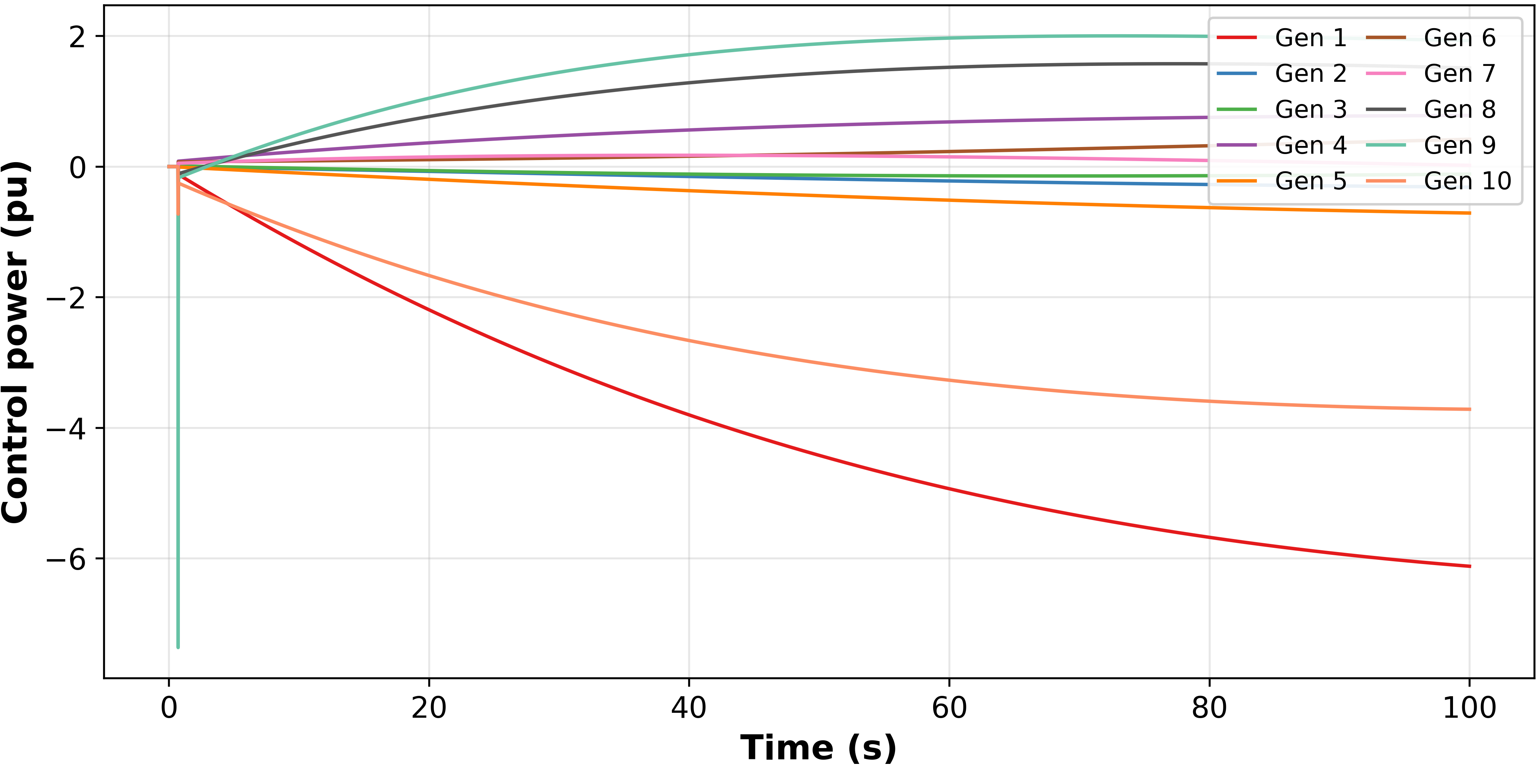}
        \caption{CPFL: control power.}
        \label{fig:f7_pu_cpfl}
    \end{subfigure}\hfill
    \begin{subfigure}{0.32\textwidth}
        \centering
        \includegraphics[width=\linewidth]{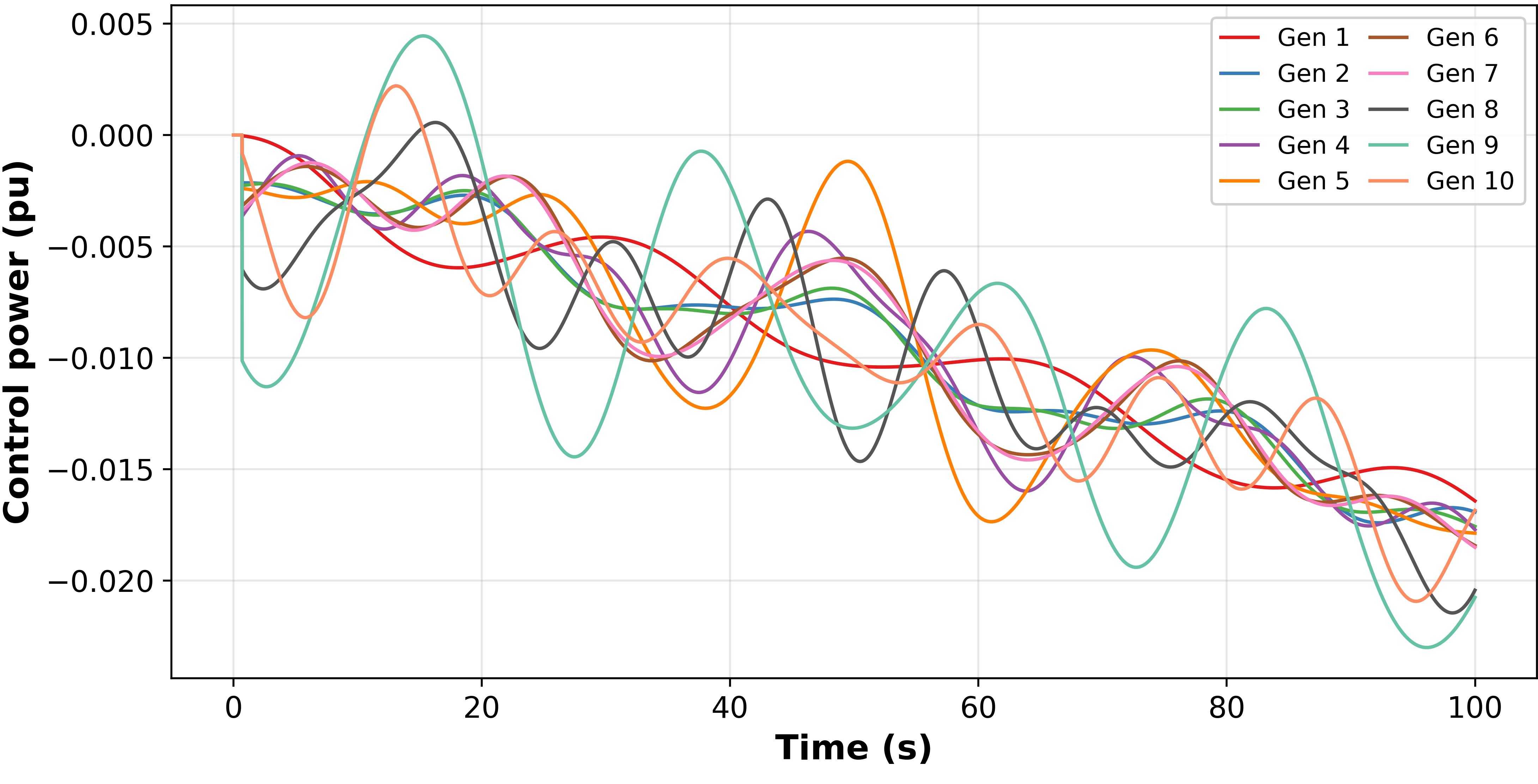}
        \caption{DPFL: control power.}
        \label{fig:f7_pu_dpfl}
    \end{subfigure}\hfill
    \begin{subfigure}{0.32\textwidth}
        \centering
        \includegraphics[width=\linewidth]{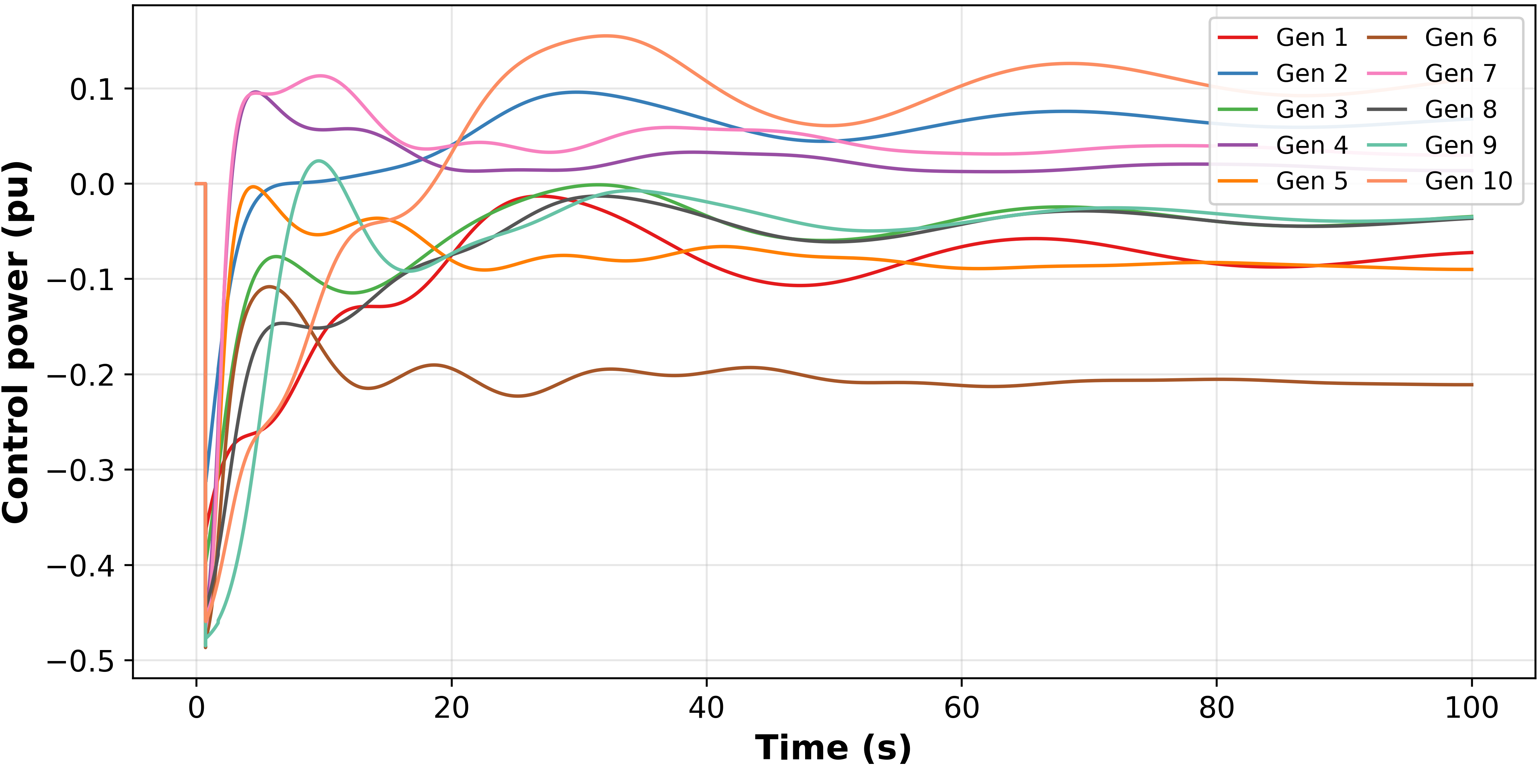}
        \caption{FedPPO-PG: control power.}
        \label{fig:f7_pu_fedppo}
    \end{subfigure}

\caption{Closed-loop frequency deviation, rotor angle, and control power trajectories under fault~F7.}
\label{fig:F7_pergen}
\end{figure*}

Fig.~\ref{fig:F7_pergen} shows the per-generator frequency deviation, rotor angle, and control power trajectories for all three controllers under the unseen fault F7 at $t_{\mathrm{cf}}{=}0.2$\,s. Fault F7 was never encountered during training, making it a direct test of out-of-distribution generalization for FedPPO-PG. FedPPO-PG (Figs.~\ref{fig:f7_omega_fedppo},~\ref{fig:f7_phase_fedppo},~\ref{fig:f7_pu_fedppo}) significantly outperforms both baselines, damping all frequency deviations to within the $\pm 0.01$\,pu stability band within approximately $6$\,s, despite never having trained on this fault location.  Rotor angles stabilize near their post-fault equilibrium values, and control power signals remain bounded within $\pm 0.5$\,pu with smooth, charging/discharging cycles.

The CPFL controller (Figs.~\ref{fig:f7_omega_cpfl},~\ref{fig:f7_phase_cpfl},~\ref{fig:f7_pu_cpfl}) achieves stabilization but at high cost: frequency deviations decay slowly over the 100,s window, with Gen~9 remaining near the $0.01$,pu threshold for an extended period, while rotor angles drift beyond $110^{\circ}$ before settling. Control power ramps continuously to over $6$,pu in Gen~1, reflecting sustained, energy-intensive injections from the linear feedback law, consistent with the mean stability time of $57.86$,s reported for F7 in Table~\ref{tab:stability}. In contrast, the DPFL controller (Figs.~\ref{fig:f7_omega_dpfl},~\ref{fig:f7_phase_dpfl},~\ref{fig:f7_pu_dpfl}) fails to stabilize within the 100,s window: inter-machine oscillations grow in both frequency and rotor angle, reaching $0.03$,pu and exceeding $120^{\circ}$, respectively. This failure stems from the lack of inter-agent coupling information in the local feedback law, preventing damping of the dominant electromechanical modes, and aligns with the $0\%$ stabilization rate in Table~\ref{tab:stability}. In contrast, FedPPO-PG maintains Gen~1 control power below $0.1$,pu after stabilization.

\subsection{Computational Efficiency}
All experiments were conducted on a Lambda workstation equipped with an AMD Ryzen Threadripper PRO 3975WX 32-core CPU, 440\,GB RAM, and dual NVIDIA RTX A4500 GPUs (20\,GB VRAM each), running Ubuntu 22.04.5~LTS. Training utilized the GPU, while inference latency was measured on CPU only to reflect realistic deployment conditions on edge hardware without dedicated accelerators.

At deployment, the FedPPO-PG controller executes as 10 independent 5-128-128-1 MLP forward passes, one per generator, totaling 345,610~FLOPs (34,561 per actor) per time step, with a mean inference time of 0.558\,ms across all 10 actors (0.056\,ms per actor) measured on CPU. This latency is well within the 10\,ms control cycle targeted for PMU-based fast controllers and significantly below the IEEE/IEC 60255-118-1-2018 reporting latency limit for P-class (Protection) devices. The standard defines the maximum allowable latency as two reporting intervals ($2/RR$); for a standard reporting rate of 60\,fps, this mandates a reporting latency of $\leq 33.33$\,ms~\cite{IEC_IEEE_60255_118_1_2018}. Consequently, the proposed controller does not introduce a computational bottleneck and is well-suited for real-time deployment in decentralized transient stability applications without requiring specialized hardware.

\subsection{Discussion}

The contrasting outcomes of CPFL and DPFL isolate the core bottleneck: not the absence of a powerful controller, but the absence of the right coupling information at the right agent. DPFL fails structurally; no gain-tuning compensates for ignoring the off-diagonal susceptance terms that dominate post-fault inter-machine dynamics, while CPFL succeeds but pays the price of sustained, energy-inefficient injections; a fixed linear gain cannot time-adapt. FedPPO-PG resolves this by replacing ad-hoc decentralization with a physics-grounded one: the $K{=}2$ neighborhood derived from post-fault $|\mathbf{B}_{i}|$ compresses the dominant coupling signal into a five-dimensional observation that is sufficient for damping and topologically consistent, which is why policies transfer to unseen faults and why FedAvg over semantically compatible actors produces a meaningful shared prior. The 72.4\% stability-time reduction and 7--14$\times$ control-effort reduction are two faces of the same learned behavior (targeted damping pulses at fault clearance tapering as the system settles), a state-dependent gain schedule no fixed linear law can replicate, and one that directly implies an order-of-magnitude reduction in ESS energy capacity requirements at deployment.

\section{Conclusion and Future Work}

This paper proposed FedPPO-PG, a federated multi-agent reinforcement learning framework for decentralized transient stability control that directly optimizes closed-loop stability objectives, addressing the robustness gap observed in prior supervised learning approaches. Three contributions drove its performance: physics-grounded neighborhood selection, which augments each agent's observation with the frequencies of its most strongly coupled electrical neighbors derived from the post-fault susceptance matrix; guided policy initialization from the classical DPFL controller; and performance-weighted federated averaging with local fine-tuning for per-generator specialization. Evaluated on the IEEE 39-bus benchmark system, FedPPO-PG achieved 100\% stabilization success-rate across all unseen fault contingencies, reducing mean stability time by 72.4\% and control effort by 7-14 times relative to the CPFL baseline, with a per-actor inference latency of 0.056\,ms and no central coordinator at deployment. Future work will investigate the optimal number of physics-coupled communication partners per agent, including learned graph-clustering approaches that jointly discover neighborhood structure and control policy via RL. Robustness to cyber-physical impairments, such as corrupted measurements, communication latency, and packet loss, represents another important open direction for practical deployment. An ablation study examining the individual contributions of each algorithmic component in FedPPO-PG will further clarify the sources of the observed performance gains.
\vspace{-1.5mm}
\section*{Acknowledgment}
The authors would like to thank Dr.\ Dileep Kalathil, Associate Professor of Electrical and Computer Engineering at Texas A\&M University, for his instruction and guidance in reinforcement learning, which supported this work.
\vspace{-1.5mm}
\bibliographystyle{IEEEtran}
\bibliography{references}

\end{document}